\documentclass[10pt,twocolumn,letterpaper]{article}

\usepackage{iccv}

\usepackage[pagebackref=true,breaklinks=true,letterpaper=true,colorlinks,bookmarks=false]{hyperref}

\iccvfinalcopy 

\def\iccvPaperID{19} 

\ificcvfinal\fi

\usepackage{times}
\usepackage{epsfig}
\usepackage{graphicx}
\usepackage{amsmath}
\usepackage{amssymb}

\usepackage{enumitem}
\usepackage[ruled]{algorithm2e}
\usepackage{array}
\usepackage{multirow}
\usepackage{color}
\usepackage{cases}
\usepackage[export]{adjustbox}
\usepackage[dvipsnames]{xcolor}
\usepackage{textpos}
\usepackage{comment}
\usepackage{textcomp}
\usepackage{listings}
\usepackage{bibentry}

\usepackage{rotating}

\usepackage{subcaption}

\usepackage{booktabs}

\usepackage[capitalize]{cleveref}
\crefname{section}{Sec.}{Secs.}
\Crefname{section}{Section}{Sections}
\Crefname{table}{Table}{Tables}
\crefname{table}{Tab.}{Tabs.}

\usepackage{soul}

\begin{document}

\title{Adaptive Self-Training for Object Detection}

\author{Renaud Vandeghen\\
{\small University of Li\`ege}\\
{\tt\small r.vandeghen@uliege.be}
\and
Gilles Louppe\\
{\small University of Li\`ege}\\
{\tt\small g.louppe@uliege.be}
\and
Marc Van Droogenbroeck\\
{\small University of Li\`ege}\\
{\tt\small m.vandroogenbroeck@uliege.be}
}
\maketitle
\ificcvfinal\thispagestyle{empty}\fi

\newcommand\blfootnote[1]{%
  \begingroup
  \renewcommand\thefootnote{}\footnote{#1}%
  \addtocounter{footnote}{-1}%
  \endgroup
}

\thispagestyle{empty}

\newcommand{\teacher}{\mathcal{T}}
\newcommand{\student}{\mathcal{S}}
\newcommand{\finetune}{\mathcal{F}}
\newcommand{\threshold}{\tau}
\newcommand{\imageset}[1]{\mathcal{D}_{#1}}
\newcommand{\red}[1]{\textcolor{red}{#1}}

\newcommand{\argmin}{\text{arg\,min}}
\newcommand{\argmax}{\text{arg\,max}}
\newcommand{\selfTrainingParameter}{\tau}
\newcommand{\datasetSymbol}{\mathcal{D}}
\newcommand{\labeledDataset}{\mathcal{D}_l}
\newcommand{\labeledDatasetStudent}{\mathcal{D}_{l'}}
\newcommand{\labeledImage}{\mathbf{I}_{i}}
\newcommand{\labeledAnnotations}{\mathbf{B}_{i}}
\newcommand{\labeledAnnotation}{\mathbf{b}_{i}^{j}}
\newcommand{\unlabeledImage}{\mathbf{U}_{i}}
\newcommand{\pseudolabeledAnnotations}{\tilde{\mathbf{B}}_{i}}
\newcommand{\unlabeledImageAfterThreshold}{\tilde{\mathbf{I}}_{i}^{\selfTrainingParameter}}
\newcommand{\pseudolabeledAnnotationsAfterThreshold}{\tilde{\mathbf{B}}_{i}^{\selfTrainingParameter}}
\newcommand{\pseudolabeledImage}{\mathbf{\tilde{b}}_{i}^{j}}
\newcommand{\pseudolabeledScore}{s_{i}^{j}}
\newcommand{\labeledDatasetWithSizeEqualTo}[1]{\datasetSymbol_{L,\,\text{size=}\text{#1}}}
\newcommand{\unlabeledDataset}{\mathcal{D}_u}
\newcommand{\candidateLabelDataset}{\mathcal{D}_{\hat{u}}}
\newcommand{\pseudoLabelDataset}{\mathcal{D}_p}
\newcommand{\trainingSetSymbol}{\mathcal{R}}
\newcommand{\validationSetSymbol}{\mathcal{V}}
\newcommand{\testSetSymbol}{\mathcal{T}}
\newcommand{\modelSymbol}{\text{M}}
\newcommand{\teacherSymbol}{\text{T}}
\newcommand{\teacherWithSize}[1]{\teacherSymbol_{#1}}
\newcommand{\studentSymbol}{\text{S}}
\newcommand{\finetuneSymbol}{\text{F}}
\newcommand{\finetuneWithSizeWithIterations}[2]{\finetuneSymbol_{#1,\,#2}}
\newcommand{\modelParameters}{\theta}
\newcommand{\meanAveragePrecisionSymbol}{\text{mAP}}
\newcommand{\intersectionOverUnionSymbol}{\text{IoU}}
\newcommand{\numN}{\mathsf{N}}
\newcommand{\numP}{\mathsf{P}}
\newcommand{\numTN}{\mathsf{T}\numN}
\newcommand{\numFP}{\mathsf{F}\numP}
\newcommand{\numFN}{\mathsf{F}\numN}
\newcommand{\numTP}{\mathsf{T}\numP}
\newcommand{\recall}{\text{R}}
\newcommand{\precision}{\text{P}}

\newcommand{\mysection}[1]{\vspace{2pt}\noindent\textbf{#1}}
\newcommand{\Table}[1]{Table~\ref{tab:#1}}
\newcommand{\Figure}[1]{Figure~\ref{fig:#1}}
\newcommand{\Equation}[1]{Equation~\eqref{eq:#1}}
\newcommand{\Equations}[2]{Equations \eqref{eq:#1} and \eqref{eq:#2}}
\newcommand{\Section}[1]{Section~\ref{sec:#1}}

\newcommand{\TODO}[1]{\textcolor{red}{[TODO:#1]}}

\newcommand{\methodName}{ASTOD\xspace}
\newcommand{\nms}{NMS\xspace}
\newcommand{\coco}{MS-COCO\xspace}
\newcommand{\dior}{DIOR\xspace}
\newcommand{\kvasir}{Kvasir-SEG\xspace}

\definecolor{myred}[a=.5]{RGB}{215,25,28} 
\definecolor{myorange}[a=.5]{RGB}{253,174,97}
\definecolor{anthoblue}[a=.5]{RGB}{31,119,180}
\definecolor{anthoorange}[a=.5]{RGB}{255,127,14}
\definecolor{anthogreen}[a=.5]{RGB}{0,150,0}
\definecolor{anthored}[a=.5]{RGB}{150,0,0}
\definecolor{anthobrown}[a=.5]{RGB}{153,76,0}
\definecolor{mygreen}[a=.5]{RGB}{166,217,106} 
\definecolor{mygray}[a=.5]{gray}{0.57}

\definecolor{newanthogreen}[a=.5]{RGB}{101,140,49}
\definecolor{newanthored}[a=.5]{RGB}{191,0,0}
\definecolor{newanthoblue}[a=.5]{RGB}{0,127,255}
\definecolor{newanthogray}[a=.5]{RGB}{76,76,76}

\definecolor{newanthoorangespotting}[a=.5]{RGB}{227,140,16}
\definecolor{newanthobluespotting}[a=.5]{RGB}{31,119,180}
\definecolor{newanthogreenspotting}[a=.5]{RGB}{44,160,44}

\definecolor{newanthoredreplay}[a=.5]{RGB}{183,27,27}
\definecolor{newanthopinkreplay}[a=.5]{RGB}{217,118,213}

\definecolor{newjacobblue}[a=.5]{RGB}{76,114,176}
\definecolor{newjacoborange}[a=.5]{RGB}{221,132,82}

\newcommand{\whitebox}{\hfill\textcolor{white}{\rule[1mm]{1.8mm}{2.8mm}}\hfill}
\newcommand{\redbox}{\hfill\textcolor{myred}{\rule[1mm]{1.8mm}{2.8mm}}\hfill}
\newcommand{\orangebox}{\hfill\textcolor{myorange}{\rule[1mm]{1.8mm}{2.8mm}}\hfill}
\newcommand{\greenbox}{\hfill\textcolor{mygreen}{\rule[1mm]{1.8mm}{2.8mm}}\hfill}
\newcommand{\graybox}{\hfill\textcolor{mygray}{\rule[1mm]{1.8mm}{2.8mm}}\hfill}
\newcommand{\BG}[1]{\textbf{{\color{red}[BG: #1]}}}

\begin{abstract}
Deep learning has emerged as an effective solution for solving the task of object detection in images but at the cost of requiring large labeled datasets. To mitigate this cost, semi-supervised object detection methods, which consist in leveraging abundant unlabeled data, have been proposed and have already shown impressive results. These methods however often rely on a thresholding mechanism to allocate pseudo-labels. This threshold value is usually determined empirically for a dataset, which is time consuming and requires a new and costly parameter search when the domain changes.
In this work, we introduce a new teacher-student method, named Adaptive Self-Training for Object Detection (ASTOD), which is simple and effective. ASTOD selects pseudo-labels  adaptively by examining the score histogram. 
In addition, we also introduce the idea to systematically refine the student, after training, with the labeled data only to improve its performance.
While the teacher and the student of ASTOD are trained separately, in the end, the refined student replaces the teacher in an iterative fashion. 

Our experiments show that, on the MS-COCO dataset, our method consistently outperforms other adaptive state-of-the-art methods, and performs equally with respect to  methods that require a manual parameter sweep search, and are therefore of limited use in practice. Additional experiments with respect to a supervised baseline on the DIOR dataset containing satellite images lead to similar conclusions, and prove that it is possible to adapt the score threshold automatically in self-training, regardless of the data distribution. The code is available at \url{https://github.com/rvandeghen/ASTOD}.
\end{abstract}

\section{Introduction}
\label{sec:Intro}

On the path to consolidate on the successes of supervised deep learning
on large labeled datasets, semi-supervised learning is
growing in interest to leverage unlabeled data and improve the performance
in many computer vision areas, when the amount of labeled data is
scarce. Particularly, semi-supervised learning has led to many improvements
for the task of image classification~\cite{Berthelot2020ReMixMatch,Berthelot2019MixMatch,Pham2021Meta,Xie2020Unsupervised,Xie2020SelfTraining,Zhai2019S4L}, and is currently growing in interest for object detection. 
According to current state-of-the-art research~\cite{Kim2022MUM,
Li2022PseCo-arxiv,
Liu2022Unbiased,
Tang2021Humble,
Wang2022OmniDETR,
Zhang2021SemiSupervised-arxiv},
semi-supervised learning methods for object detection (SSOD) are usually based on the principle of self-training, wherein a teacher model is first trained with the labeled data in order to generate pseudo-labels for unlabeled data. Then a second model, called the student, is trained with the pseudo-labeled data. Most of the time, the teacher and the student are trained at the same time in a mutual way.

How can we effectively endorse candidate labels generated by methods in the context of SSOD? This question becomes particularly important when considering state-of-the-art classification methods applied to object detection tasks.
More precisely, one has to answer the question of how far endogenous
(candidate) labels created by a teacher are to be trusted so that,
when added to the labeled dataset, the detection performance of a student
network twinned with the teacher network can be improved. Keeping
only trustable labels by thresholding the predictions provided
by the teacher based on their confidence scores is a simple yet effective method. But beyond this simplicity,
determining the adequate threshold value remains tricky. Current works in SSOD often require a costly parameter sweep across different values to determine a suitable threshold. While it is easy to understand the behavior of such a threshold regarding the generation of false positives or false negatives, it is not clear which threshold to choose, as evidenced by previous works where the reported optimal threshold value ranges between $0.5$ and $0.9$ depending on the datasets and network architectures. Also, most works only cover the case
of natural scenes, such as \coco~\cite{Lin2014Microsoft} and PASCAL VOC~\cite{Everingham2010PascalVOC},
preventing drawing conclusions for other datasets.


In this paper, we introduce our Adaptive Self-Training for Object Detection method (\methodName) to perform the task of object detection. The main idea of our method is to determine the threshold value applied to the confidence score to select pseudo-labels adaptively which is based on the score histogram of the pseudo-labels. In addition, this strategy has the benefit of determining a threshold value for each class without additional cost, which would be very costly with a parameter sweep for most practical semi-supervised setups. On top of this adaptive threshold, we use different views of the unlabeled images during the generation of pseudo-labels to improve the predictions of the teacher by reducing the number of missed objects, and to improve the predictions of the bounding box coordinates. It is also important to account for the uncertainty in the pseudo-labeled data when we use them. To do so, we downscale the contribution of pseudo-labels in the loss based on their confidence scores. Lastly, we refine our student with the labeled dataset before using it as our new teacher in an iterative way. 

\begin{figure*}
    \centering
    \includegraphics[width=\linewidth]{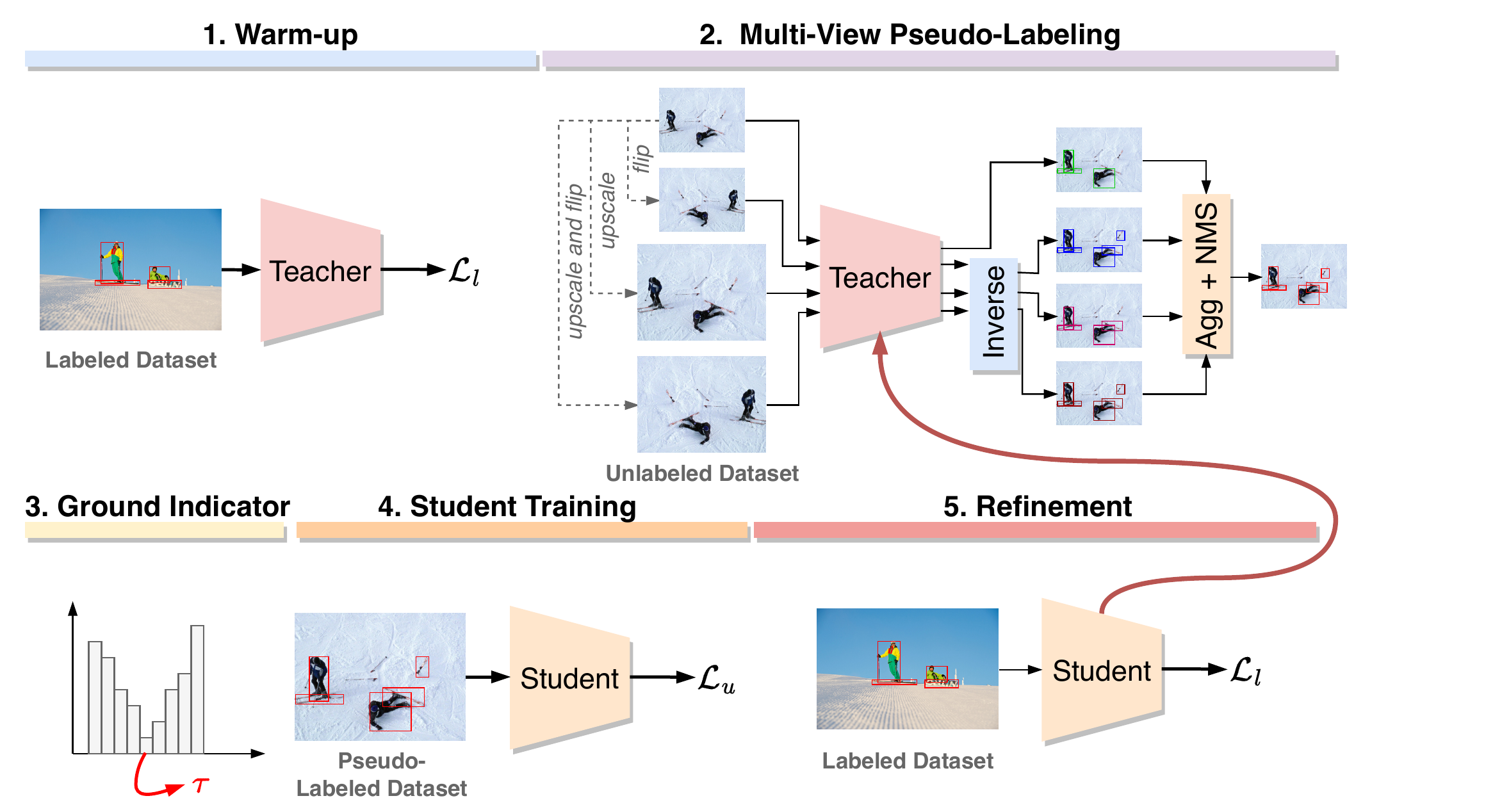}
    \caption{Pipeline of our self-training \methodName method. (1) A teacher is trained with the labeled dataset. (2) We use the teacher to generate candidate labels on the unlabeled data using multiple views. We apply the inverse view transformation to gather the different predictions in the same dimensional space. The predictions are then merged with NMS. (3) Based on the confidence score histogram, we determine the threshold value $\selfTrainingParameter$ to filter the candidate boxes, leading to a pseudo-labeled dataset. (4) Next, we train the student with the labeled and pseudo-labeled datasets. (5) Finally, we refine the student with the labeled dataset and use it to replace the teacher. ASTOD can then be used in an iterative fashion by replacing the teacher (2) with the refined student.}
    \label{fig:Drawing-showing-the-five-steps}
\end{figure*}

In Section \Section{Method}, we delve into the details of our \methodName
method, whose pipeline is illustrated in \Figure{Drawing-showing-the-five-steps}, after a formal definition of our problem statement. Later, in \Section{Exp}, we validate the principle of adaptive self-training with \methodName 
for two experimental setups, namely \textit{COCO-standard} and \textit{DIOR}. 

Our contributions can be summarized as follows.
\begin{itemize}
\item  We present a novel end-to-end SSOD method, called \methodName, based on an iterative teacher-student framework. This method includes a computational-free heuristic based on the score histogram to determine the threshold value for the selection of pseudo-labels.
\item We show that using multiple views to generate candidate labels is a simple yet effective technique to improve the labeling process. 
\item We show that the systematic use of a refinement step is crucial to improve the performance of the student.
\item We demonstrate its effectiveness against state-of-the-art methods for two  setups.
\end{itemize}

\section{Related Work}
\label{sec:SOTA}

\mysection{Semi-supervised learning.}
Semi-supervised learning has already been thoroughly studied for image classification. Among the achievements, some methods are
based on the principles of consistency training~\cite{Bachman2014Learning,Berthelot2020ReMixMatch,Berthelot2019MixMatch,Laine2016Temporal-arxiv,Xie2020Unsupervised,Zhai2019S4L},
which forces the invariance of a model with respect to input noise
by introducing a regularization loss for the unlabeled data. For example, Zhai
\etal~\cite{Zhai2019Adversarially-arxiv} used consistency training to improve the
robustness of the model under adversarial attacks. Xie \etal~\cite{Xie2020Unsupervised}
have tried another approach in which they minimize the divergence
between the output prediction of an unlabeled image and
its augmented counterpart. 

Another principle for semi-supervised learning
is self training~\cite{Lee2013Pseudo,Li2020Improving,Pham2021Meta-arxiv,Rosenberg2005SemiSupervised,Sohn2020ASimple-arxiv,Xie2020SelfTraining}.
It consists of three parts. First, a teacher
model is trained with the labeled data. Then, the trained teacher
model is used to generate pseudo-labels on the unlabeled data. Finally,
a student model is trained with a dataset comprising the original
labels and the pseudo-labels. In particular, Xie \etal~\cite{Xie2020SelfTraining}
showed that adding noise during the training of the student model
and increasing the network capacity lead to state-of-the-art results.

\mysection{Semi-supervised object detection.}
Driven by the successes obtained for image classification, different
semi-supervised learning methods have been tailored for the specific
task of the object detection~\cite{Jeong2019Consistency,
Kim2022MUM,
Li2020Improving,
Liu2021Unbiased,
Liu2022Unbiased,
Li2022PseCo-arxiv,
Sohn2020ASimple-arxiv,
Tang2021Humble,
Wang2022OmniDETR,
Zhang2021SemiSupervised-arxiv,
Zoph2020Rethinking},
even though pioneering work began in 2005~\cite{Rosenberg2005SemiSupervised}.
Among them, most methods~\cite{
Kim2022MUM,
Li2022PseCo-arxiv,
Li2020Improving,
Liu2021Unbiased,
Liu2022Unbiased,
Sohn2020ASimple-arxiv,
Wang2022OmniDETR,
Zhang2021SemiSupervised-arxiv} use a threshold value determined empirically to select or reject a pseudo-label. Only few of them are designed without threshold~\cite{Jeong2019Consistency, Tang2021Humble}.
Particularly, one of the first work was done by Jeong \etal~\cite{Jeong2019Consistency}, who proposed a consistency-based semi-supervised learning method by applying consistency between
a horizontally flipped image and the original one for the classification
part, with the Jensen-Shannon divergence, as well as for the localization
part, with a weighted sum of the squared errors of the four different
components of the localization loss. They applied this consistency
loss for labeled and unlabeled data. Given that this loss can
be dominated by the background class, they performed a background
elimination, which removes predictions likely to belong to the background.
Another approach is to use soft pseudo-labels~\cite{Tang2021Humble}, which means
that the whole distribution of class probabilities is used rather
than the hard pseudo-label. Those methods give more flexibility as they do not need any threshold value. Among the threshold-based methods, the field of SSOD has grown in interest after that
Sohn \etal~\cite{Sohn2020ASimple-arxiv} introduced a semi-supervised
learning method based on self-training and augmentation-driven
consistency regularization. They start by training a teacher in a
supervised manner. Afterwards, they use the teacher to generate candidate labels, which are selected when their confidence scores are above a threshold of $0.9$. The method then uses a second
model, which is trained on both the labeled and pseudo-labeled data
by jointly minimizing a conventional supervised loss and a weighted
unsupervised loss based on consistency regularization with strong
data augmentations. In~\cite{Li2020Improving}, Li \etal presented
a selective self-supervised self-training for object detection method.
They started with a teacher-student self-training method with a threshold
value of $0.7$ during pseudo-labeling, and improved the pseudo-labeling
step with a so-called selective network. This network splits the set
of pseudo-labels into three categories (positive, negative, and ambiguity),
but only the positive pseudo-labels are considered in the loss term.
They also implemented a consistency term in their loss based on the
work presented in~\cite{Jeong2019Consistency}, which is also only
used for the positive pseudo-labels. Liu \etal~\cite{Liu2021Unbiased}
proposed an unbiased teacher, which is a teacher-student method trained
in a mutual setting. The teacher generates pseudo-labels for the training
of the student and, then, the teacher is updated with exponential moving
average (EMA), leading to continually improving models. The authors
also used a threshold value of $0.7$ to remove boxes with low confidence
scores and they addressed the problem of class imbalance by replacing
the standard cross-entropy loss with the focal loss~\cite{Lin2017Focal}. They also published a second version of their unbiased teacher~\cite{Liu2022Unbiased} for anchor-free detectors. Zhang \etal~\cite{Zhang2021SemiSupervised-arxiv}
also addressed the problem of class imbalance with two modules. The
first module addresses the problem of foreground-background imbalance
by pasting synthetic objects from the training/pseudo dataset in the
training images. The second module addresses the problem of foreground-foreground
imbalance, which changes the sampling probability with respect to
the class occurrence. Kim \etal~\cite{Kim2022MUM} presented a data augmentation technique for the unlabeled dataset that mixes image tiles and feature tiles together and then unmixes the features for the student. Those unmixed feature maps are then processed by the RPN and ROI heads with the pseudo-labels being generated by the teacher with the same weakly augmented images. Li \etal~\cite{Li2022PseCo-arxiv} also adopted the teacher-student dual learning but took into account the noisy nature of pseudo-boxes regression. Their method is based on a learning scheme that uses multiple views for both the images and the feature maps to enforce consistency.
Tanaka \etal~\cite{Tanaka2022NonIterative} recently proposed to optimize the threshold based on the $\beta$-score and without iterating on the student.\newline
The current literature on the topic of semi-supervised learning for object detection exhibits a wide variety of nuances around a teacher-student scheme and the calculation of pseudo-labels. Alongside, this often results in the heuristic determination of parameters among which the determination of a threshold on the confidence score during the generation of pseudo-labels. As opposed to most approaches of current literature, we intentionally skip this process thanks to an adaptive calculation of such threshold embedded into a new iterative, multiple-view, and refined teacher-student scheme. This forms the basis of our concept of adaptive self-training. In the case of \methodName, this calculation occurs by analyzing the histogram of scores associated to the generation of pseudo-labels in a fully automatic fashion. 

\section{Method}
\label{sec:Method}

\mysection{Problem statement.}
%
We consider a set, $\datasetSymbol$, of images, $\boldsymbol{x}_i$, containing
several classes of objects to be detected.
Among $\datasetSymbol$, only a subset of images, ${\boldsymbol{x}_i^l}$, are annotated with the class and localization of all objects of interest, $\boldsymbol{y}_i^l$
(called “labels” or ground-truths), and compose the subset
of labeled images $\labeledDataset = \{\boldsymbol{x}_i^l, \boldsymbol{y}_i^l \}_{i=1}^{N_l}$, where $N_l$ is the number of labeled images in $\datasetSymbol$. Each ground-truth $\boldsymbol{y}_i^l$ is composed by a set of classes $\boldsymbol{c}$ and bounding box coordinates $\boldsymbol{b}$. The remaining images of $\datasetSymbol$ with no labels, $\boldsymbol{x}_i^u$,  compose the subset of unlabeled images $\unlabeledDataset = \{\boldsymbol{x}_i^u \}_{i=1}^{N_u}$, where  $N_u$ is the number of unlabeled images in $\datasetSymbol$. These subsets are complementary sets (that is $\datasetSymbol=\labeledDataset\,\cup\,\unlabeledDataset$), and we assume that they come from the same data distribution. In semi-supervised learning setups, we often have $N_l \ll N_u$.

\mysection{Teacher warm up.}
Our method relies on a teacher-student scheme, where the student learns from the pseudo-labels generated by the teacher. Thus, the first step is to learn a teacher that is able to generate high-quality candidate labels, which are all the predictions made by a model without restrictions. The first step of our method then consists in training the teacher model with the labeled data only.
We use the conventional training loss for object detection, which is the sum of the classification and regression losses:
\begin{equation}\label{eq:loss_s}
    \mathcal{L}_l = \sum_{i=1} \left[\sum_{j=i} \left(\mathcal{L}_{cls}\left(p(c_j | \boldsymbol{x}_i), c_j\right)
     +  
    \mathcal{L}_{reg}\left(p(\boldsymbol{b}_j | \boldsymbol{x}_i), \boldsymbol{b}_j\right)
    \right) \right],
\end{equation}
where the index $j$ corresponds to the index of an anchor, $p(c_j | \boldsymbol{x}_i)$ is the predicted class probability of anchor $j$ in the image $\boldsymbol{x}_i$, and $p(\boldsymbol{b}_j | \boldsymbol{x}_i)$ are the 4 bounding box coordinates of a predicted bounding box.

\mysection{Multi-view pseudo-labeling and ground threshold.}
After we warm up the teacher model, we use it to generate candidate labels for the unlabeled data. Since we want to mitigate the false negatives due to missed predictions and we want the most accurate predictions, the inference of each unlabeled image is processed under multiple views: original image, horizontally flipped image, rescaled image, and both flipped and rescaled image. Afterwards, we apply the inverse transformations to the predictions so that they can be aggregated in the same dimensional space. To reduce redundancy, we apply non-maximum suppression (\nms) on each prediction before and after aggregation. This leads to a subset of candidate pseudo-labels $\candidateLabelDataset = \{\boldsymbol{x}_i^u, \hat{\boldsymbol{y}}_i^u \}_{i=1}^{N_c}$, with $N_c$ being the number of unlabeled images which have at least one prediction $\hat{\boldsymbol{y}}$, given that we automatically discard images without prediction. Note that each prediction $\boldsymbol{\hat{y}}_i^u$ is composed by the set of predicted classes, its corresponding bounding box coordinates, and the confidence scores $\boldsymbol{s}$ associated to each box.
Now that we have access to high-quality candidate labels, we need to select among them those that can be considered as true positives. In contrast to classification tasks, where we can select the class with the highest probability, this is a challenging step for SSOD. Indeed, there can be multiple objects in the same image, meaning that an independent decision must be taken for each anchor. The most straightforward and, by far, most common solution is to threshold the candidates based on their score predictions.

Previous works in the field use a threshold value, denoted by $\selfTrainingParameter$, which suits at best their method and the dataset on which they evaluate it. Typical values for this threshold range between $0.5$ and $0.9$. However, this threshold value is often determined with a costly parameter sweep, unique for all classes, and is optimized for only one image distribution (natural scenes with \coco).
To account for those shortcomings, we propose a new heuristic, called \emph{ground thresholding}, based on the score histogram to determine the threshold value: ground thresholding selects the bin with the lowest density. From our experience, taking the bin with the lowest density is a heuristic that constitutes a well-suited compromise solution for eliminating false positives (bins on the left) while preserving a high enough recall (bins on the right). The final pseudo-labeled dataset is then $\pseudoLabelDataset = \{\boldsymbol{x}_i^p, \hat{\boldsymbol{y}}_i^p \}_{i=1}^{N_p}$, with $N_p$ being the number of candidate images which have at least one prediction that satisfies $\hat{\boldsymbol{y}}^c \geq \selfTrainingParameter$.
Since this heuristic does not require the burden of a parameter sweep to find the threshold value, it can be applied independently for each class, which does not bias the threshold value with respect to the class occurrence. It also means that our method can easily generalize to any dataset without any additional computational cost.

\mysection{Iterative student training.}
We train the student model in the same manner as for the teacher, but with the labeled and pseudo-labeled data ($\labeledDatasetStudent = \labeledDataset \cup \pseudoLabelDataset$). During the training of the student, we do not distinguish images coming from $\labeledDataset$ or $\pseudoLabelDataset$. However, to account for the uncertainty in the pseudo-labeled data, we generalize the weighting term
\begin{equation}\label{eq:alpha_sst}
    \alpha_j =
    \left\{
    \begin{array}{ll}
       \frac{s_j-\tau_l}{\tau_h-\tau_l} & \text{if \;  $\threshold_l \leq s_j < \threshold_h$},\\
       1 & \text{otherwise.}
    \end{array}
  \right.
\end{equation}
used in~\cite{Vandeghen2022SemiSupervised} for the loss by fixing $\threshold_h = 1$, where $\tau_h$ and $\tau_l$ represent a high and a low threshold value and $s_j$ the score prediction. This leads to the weighted loss function:
\begin{equation}\label{eq:loss_u}
    \mathcal{L}_u = \sum_{i=1} \left[\sum_{j=i} \alpha_j \left(\mathcal{L}_{cls}
     +  
    \mathcal{L}_{reg}
    \right) \right],
\end{equation}
where $\mathcal{L}_{cls}$ and $\mathcal{L}_{cls}$ are the same classification and regression losses as in \Equation{loss_s}. The weighting factor $\alpha_j$ used to reduce the contribution of each prediction is then defined as
\begin{equation}\label{eq:alpha}
    \alpha_j =
    \left\{
    \begin{array}{ll}
       \frac{s_j-\threshold_j}{1-\threshold_j} & \text{if \;  $\threshold_j \leq s_j \leq 1$},\\
       1 & \text{otherwise,}
    \end{array}
  \right.
\end{equation}
with $s_j$ and $\threshold_j$ being the score and the class-wise threshold value associated to the prediction. Since the score value of labeled data are implicitly set to $1$, only pseudo-labeled data contribute to the weighting factor of the loss.

The final step consists in the refinement of the trained student model with the labeled data only. Our method can then be used in an iterative pipeline, where the refined student model will become the new teacher. Since we expect that the student model achieves better results compared to the teacher, its predictions for the candidate labels should be of higher quality which thus leading to an even better new student.
\section{Experiments}
\label{sec:Exp}

\subsection{Experimental setup}
\mysection{Datasets.}
Our experimental setup follows the methodology introduced in STAC~\cite{Sohn2020ASimple-arxiv}. In particular, we evaluate our method on two setups: (\emph{setup} 1) natural images on \coco~\cite{Lin2014Microsoft} and (\emph{setup} 2) satellite images from the \dior~\cite{Li2020Object} dataset. For the first setup (called \textit{COCO-standard} hereafter), we randomly sample $1, 2, 5$ and $10\%$ labeled 
training data out of the $118$k images available in the \texttt{train2017} split and use the remaining ones as unlabeled training data. For the second setup (\textit{\dior}), we first shuffle all the labeled images in two parts: the training part with $80\%$ and the validation part with the remaining $20\%$. Then, we sample $10\%$ of the training dataset as labeled data and the remaining $90\%$ as unlabeled data. 
Unlike most of the other works in semi-supervised for object detection, which use PASCAL VOC~\cite{Everingham2010PascalVOC} as second dataset, we evaluate our method on satellite images with the \dior dataset to analyze our method for a totally different image distribution.\\
For both \textit{COCO-standard} and \textit{\dior}, we report the mean and standard deviation of the $AP_{50:95}$ ($\meanAveragePrecisionSymbol$) over $5$ folds.

\mysection{Implementation details.}
For a fair comparison with previous works, we use Faster-RCNN~\cite{Ren2017Faster} with FPN~\cite{Lin2017Feature} and a ResNet-50~\cite{He2016DeepResidual} backbone pretrained on ImageNet~\cite{Russakovsky2015ImageNet} as object detector. For the teacher warm-up, we train the model for $20$k steps of gradient descent with a starting learning rate of $0.08$ that decays after $13$k and $18$k steps by a factor 10. For the generation of pseudo-labels, we use $4$ different views of the unlabeled image: $(1)$ normal view, $(2)$ upscale of the original image by a factor of $2$, $(3)$ horizontal flip of the original image, and $(4)$ both upscaling and flipping of the original image. From the score histogram, we set the threshold by selecting the bin with the lowest density between $0.5$ and $1$ with $21$ bins ---the choice of the $[0.5, 1]$ range was motivated by the need to select only pseudo-labels with enough confidence, while we choose $21$ bins because we wanted an odd number of bins and, by experience, the impact of more bins on the threshold value was insignificant. The student models are trained for $180$k steps, with the same learning rate as the teacher, which follows the same decay strategy after $120$k and $160$k steps. Finally, the student models are refined on the labeled dataset for $10$k steps with a learning rate starting at $0.0008$ which decays after $6$k steps. All the models for \textit{COCO-standard} are trained on $4$ GPUs, with a batch size of $16$ per GPU. We apply random color and scale jitter as data augmentation. When we train the student models, the batches are formed with $2$ labeled and $14$ unlabeled images. For \textit{\dior}, we use $3$ scale levels per anchor to better match the different bounding box shapes. The batch size is reduced to $8$ per GPU and the student is trained for $90k$ steps.

\subsection{Results}

\mysection{COCO-standard.}
We compare our model with the state-of-the-art semi-supervised object detection methods on \textit{COCO-standard}, as it is the main benchmark adopted by the SSOD community. We group the different methods according to how their threshold value is set, if any. In particular, we group methods that perform a parameter search to find the optimal threshold value. This kind of methods represents most of previous works in the field, such as STAC~\cite{Sohn2020ASimple-arxiv}, Soft Teacher~\cite{Xu2021Soft} or Unbiased Teacher~\cite{Liu2021Unbiased, Liu2022Unbiased}. The second group is for methods that do not need an empirical search for their threshold, such as CSD~\cite{Jeong2019Consistency}, Humble Teacher~\cite{Tang2021Humble} and our method. One could say that the former group are dataset dependent while the latter ignore the dataset distribution. Even though our \methodName method has a threshold parameter, it is adaptive to the dataset, thus closer to the second group than the first. The results are shown in \Table{coco_comparison}, where our results are obtained after $3$ iterations of student training plus refinement. While being competitive \wrt to the state-of-the-art methods with empirical threshold, like Unbiased Teacher v2~\cite{Liu2022Unbiased} and PseCo~\cite{Li2022PseCo-arxiv}, our method consistently outperforms methods that do not take into account the dataset distribution~\cite{Jeong2019Consistency, Tang2021Humble}. It is important to note that if Unbiased Teacher v2~\cite{Liu2022Unbiased} and PseCo~\cite{Li2022PseCo-arxiv} have better performances than \methodName, they would be more challenging to use in practice on a new dataset, simply because there is no data to fine-tune their thresholds.

\begin{table*}[t]
\centering
\caption{Experimental results on \textit{COCO-standard} for the mAP: we report the mean and standard deviation over $5$ randomly sampled dataset. We group the different methods \wrt to their thresholding strategy. The methods in the middle of the table use a manual  empirical search for the threshold value (these methods are thus intractable when applied on a new unknown  domain), while methods in the lower part are fully automatic. The results of Supervised$\dagger$ represents the performance of our teacher, which sets our supervised baseline. The results of our method are obtained after $3$ iterations of student with refined models, where we used our ground threshold and the multi-views strategies during the pseudo-labeling step.} 
\label{tab:coco_comparison}
\resizebox{0.8\linewidth}{!}{%
\begin{tabular}{ccccc}
\toprule
  &\multicolumn{1}{c}{1\%}     & \multicolumn{1}{c}{2\%}    & \multicolumn{1}{c}{5\%}    & \multicolumn{1}{c}{10\%}     \\ \hline
Supervised & $9.05 \pm 0.16$      &    $12.70 \pm 0.15$    &    $18.47 \pm 0.22$    &    $23.86 \pm 0.81$ \\
Supervised$\dagger$ & $12.14 \pm 0.21$      &    $16.67 \pm 0.30$    &    $23.59 \pm 0.20$    &    $29.34 \pm 0.20$ \\
\hline
STAC~\cite{Sohn2020ASimple-arxiv}     & $13.97 \pm 0.35$ & $18.25 \pm 0.25$ & $24.38 \pm 0.12$ & $28.64 \pm 0.21$ \\ 
Instant Teaching~\cite{Zhou2021InstantTeaching}     & $18.05 \pm 0.15$ & $22.45 \pm 0.15$ & $26.75 \pm 0.05$ & $30.40 \pm 0.05$ \\
ISMT~\cite{Yang2021Interactive}     & $18.88 \pm 0.74$ & $22.43 \pm 0.56$ & $26.37 \pm 0.24$ & $30.53 \pm 0.52$ \\
Unbiased Teacher~\cite{Liu2021Unbiased} & $20.75 \pm 0.12$ & $24.30 \pm 0.07$ & $28.27 \pm 0.11$ & $31.50 \pm 0.10$ \\ 

Soft Teacher~\cite{Xu2021Soft}     & $20.46 \pm 0.39$ & - & $30.74 \pm 0.08$ & $34.04 \pm 0.14$ \\
Omni-DETR~\cite{Wang2022OmniDETR} & $18.6$ & $23.2$ & $30.2$ & $34.1$ \\
Unbiased Teacher v2~\cite{Liu2022Unbiased} & $\mathbf{25.40 \pm 0.36}$ & $\mathbf{28.37 \pm 0.03}$ & $31.85 \pm 0.09$ & $35.05 \pm 0.02$ \\
PseCo~\cite{Li2022PseCo-arxiv} & $22.43 \pm 0.36$ & $27.77 \pm 0.18$ & $\mathbf{32.50 \pm 0.08}$ & $\mathbf{36.06 \pm 0.24}$ \\
\hline
CSD~\cite{Jeong2019Consistency}    & $10.51 \pm 0.06$ & $13.93 \pm 0.12$ & $18.63 \pm 0.07$ & $22.46 \pm 0.08$ \\
Humble Teacher~\cite{Tang2021Humble}   & $16.96 \pm 0.38$ & $21.72 \pm 0.24$ & $27.70 \pm 0.15$ & $31.61 \pm 0.28$ \\
\textbf{Ours (\methodName)} & $\mathbf{19.47 \pm 0.39}$ & $\mathbf{24.85 \pm 0.21}$ & $\mathbf{30.43 \pm 0.50}$ & $\mathbf{34.58 \pm 0.22}$ \\
\bottomrule
\end{tabular}}
\end{table*}

\mysection{DIOR.} It is important to design SSOD methods that are usable and effective in many different setups. While previous works in the field have mainly focused on natural scene images with \coco and PASCAL-VOC, we decided to evaluate our method with a different and challenging setup. We targeted the field of satellite images because of the growing interest around it, with the \dior dataset.

Since there is no baseline for SSOD methods for that dataset, we will compare with the supervised baseline achieved by our teacher model. We generated candidate labels with the Flip+Scale strategy and used a class-wise ground threshold. We also refine the student model to further improve its performance. The results of the different models are presented in \Table{ablation_dior}.
The gain obtained after one iteration shows the effectiveness and robustness of our method towards completely different data distribution, meaning that it can be further used in other applications. 

\begin{table}[t]
\centering
\caption{Comparison between models trained on \textit{DIOR}. The refined student models are trained with candidates labels generated with the Scale+Flip technique and a class-wise threshold. We report the mean and standard deviation over 5 randomly sampled dataset.
    }
\label{tab:ablation_dior}
\begin{tabular}{cccc}
\toprule
  & Supervised & Student & Refined \\
\hline
mAP & $47.59 \pm 0.36$ & $51.23 \pm 0.35$ & $\mathbf{52.89 \pm 0.33}$ \\

\bottomrule
\end{tabular}
\end{table}

\subsection{Ablation study}\label{sec:ablation}

We study our method \wrt to its different components on \textit{COCO-standard} with a labeled dataset size of $10\%$.

\begin{figure*}
    \centering
         \begin{subfigure}[b]{0.18\textwidth}
         \centering
         \includegraphics[width=\textwidth]{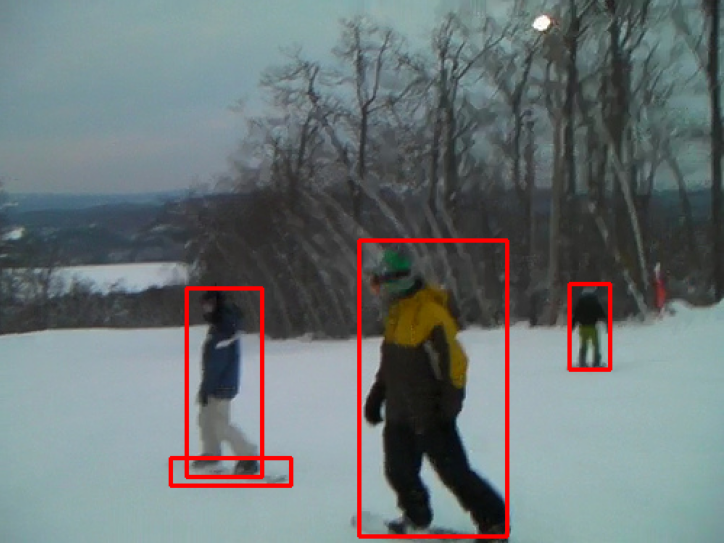}
         \caption{Normal view}
         \label{fig:normal}
     \end{subfigure}
     \hfill
         \begin{subfigure}[b]{0.18\textwidth}
         \centering
         \includegraphics[width=\textwidth]{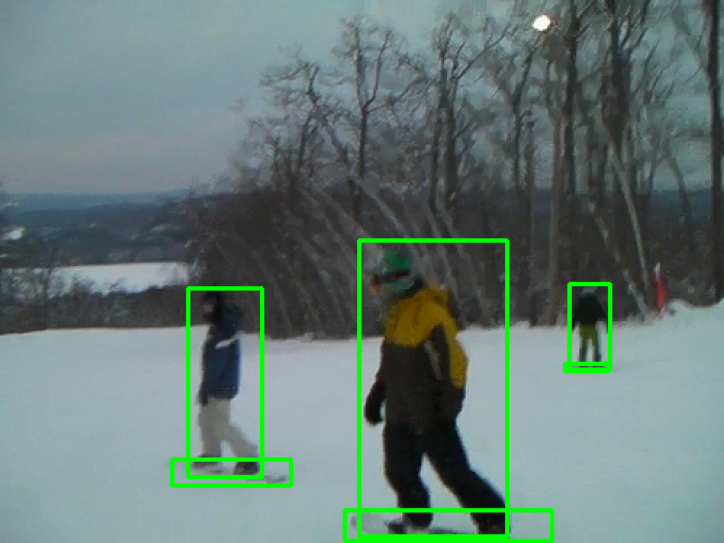}
         \caption{Scaled view}
         \label{fig:scale}
     \end{subfigure}
     \hfill
         \begin{subfigure}[b]{0.18\textwidth}
         \centering
         \includegraphics[width=\textwidth]{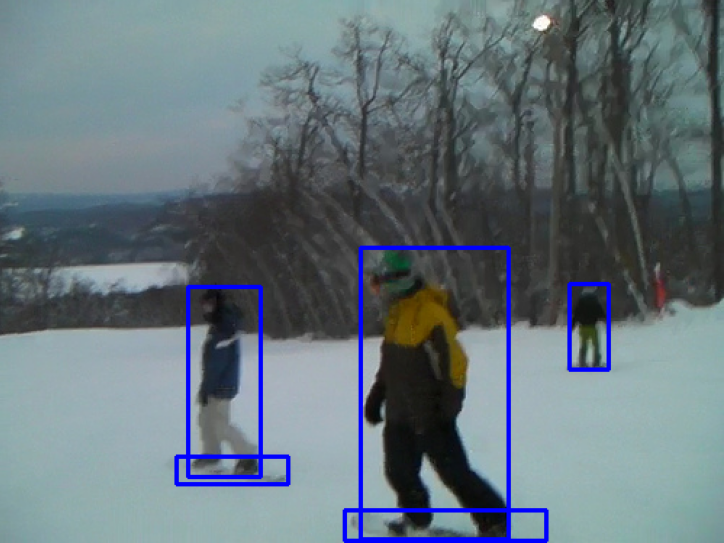}
         \caption{Flipped view}
         \label{fig:flip}
     \end{subfigure}
     \hfill
         \begin{subfigure}[b]{0.18\textwidth}
         \centering
         \includegraphics[width=\textwidth]{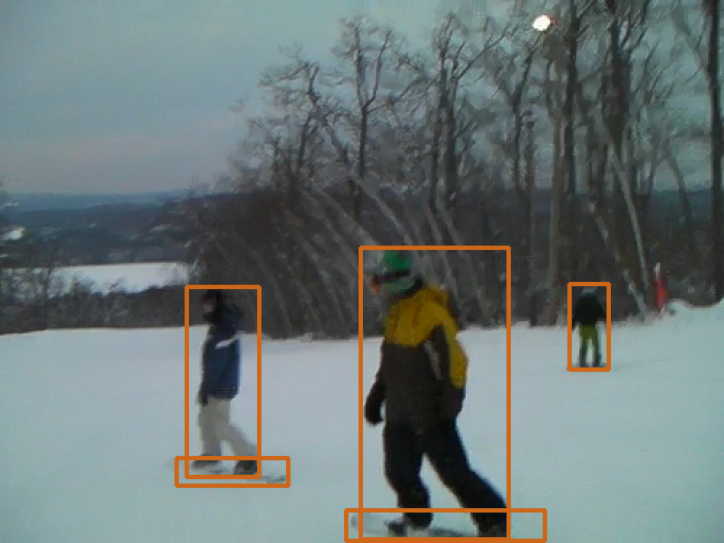}
         \caption{Scaled+Flipped view}
         \label{fig:scale_flip}
     \end{subfigure}
     \hfill
         \begin{subfigure}[b]{0.18\textwidth}
         \centering
         \includegraphics[width=\textwidth]{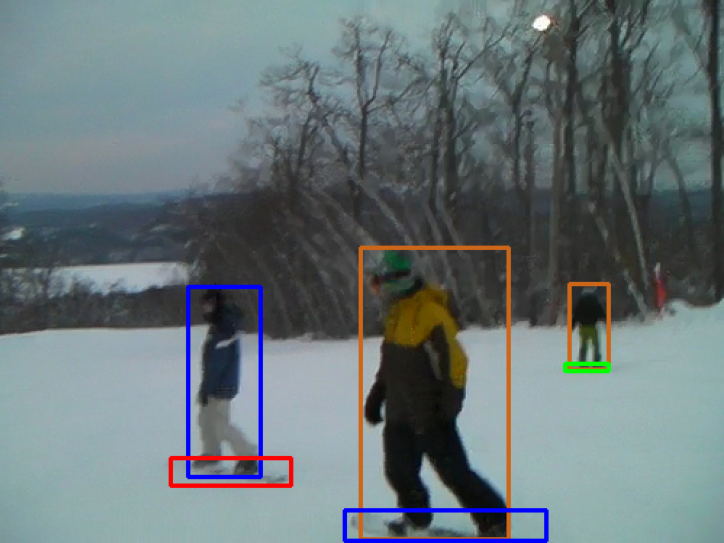}
         \caption{Aggregated view}
         \label{fig:aggregated}
     \end{subfigure}
    \caption{Comparison between the candidate labels for the different views. The normal view (a) misses two snowboards. Both flipped and scaled+flipped views (c) and (d) miss the small snowboard. Only the scaled view (b) has detected all the snowboards. The aggregated view (e) combines the information of all images (with NMS) to produce the final candidate labels. Note that images (b), (c) and (d) are transformed back to the original space.}
    \label{fig:Multiview}
\end{figure*}

\mysection{Pseudo-labeling.}
It is important to rely on high-quality pseudo-labels. To obtain those high-quality pseudo-labels, it is possible to use a high threshold value but at the cost of rejecting potential true positives with lower confidence scores. However, it is not possible to avoid false negatives due to missed predictions. Our multi-view pseudo-labeling strategy can help to reduce their numbers. \Figure{Multiview} shows the predicted candidate labels for the different views we consider. We can effectively see that only using the normal view fails to predict some objects in the image, such as the right snowboard, and adding the predictions of the other views solves the problem. Since our aggregation of boxes is performed with NMS, which is a score-based method, the final boxes are a mix of the different views, leading to the best possible candidates. 

\begin{figure}[ht]
    \centering
         \begin{subfigure}{.49\linewidth}
  \centering
  \includegraphics[width=\linewidth]{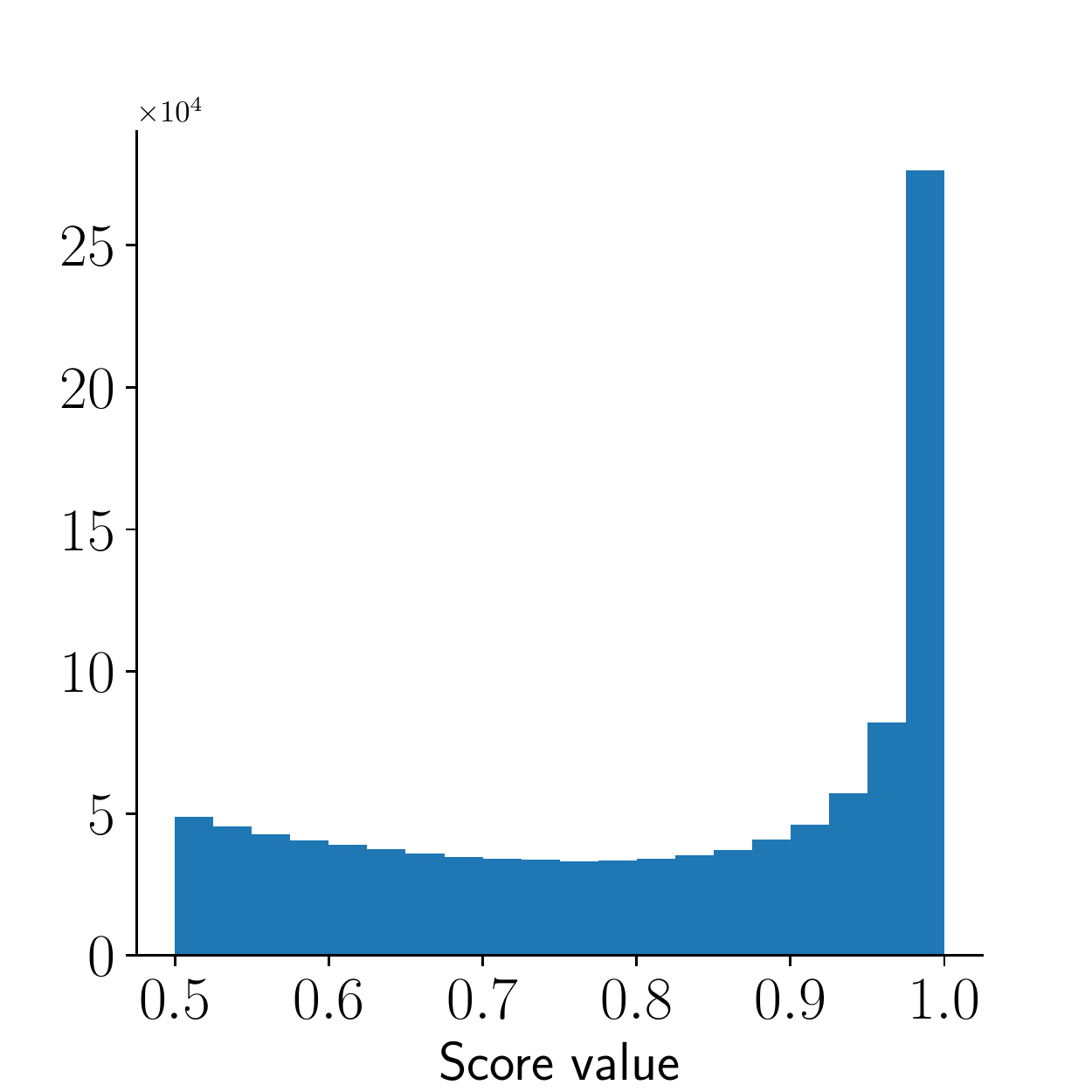}
  \caption{Histogram ranging from $0.5$ to $1$ with $21$ bins.}
  \label{fig:histo_1}
\end{subfigure}%
\hfill
\begin{subfigure}{.49\linewidth}
  \centering
  \includegraphics[width=\linewidth]{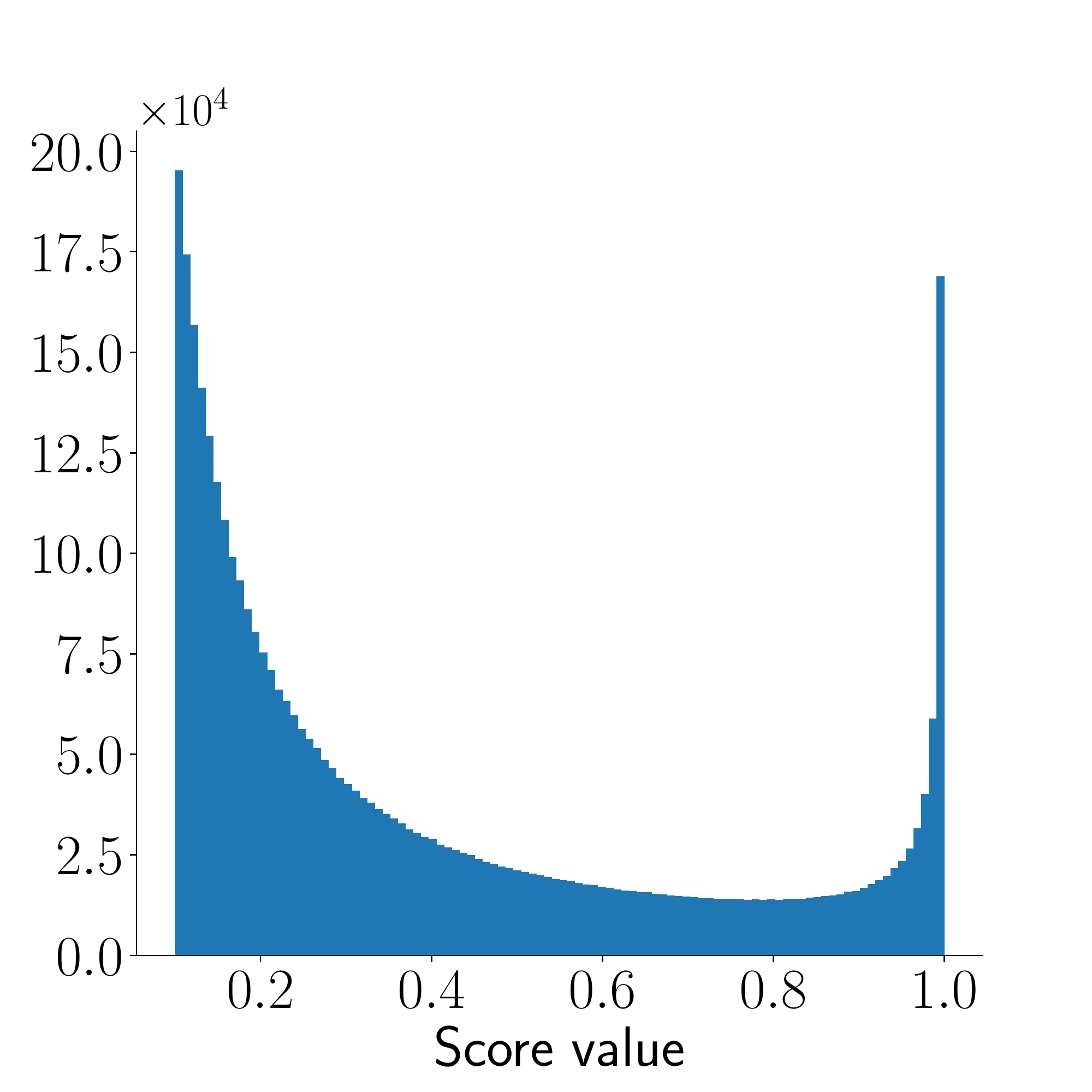}
  \caption{Histogram ranging from $0.1$ to $1$ with $101$ bins.}
  \label{fig:histo_2}
\end{subfigure}
    \caption{Histograms for different parameters.}
    \label{fig:histogram_comparison}
\end{figure}

\begin{figure}
    \centering
         \begin{subfigure}{.49\linewidth}
  \centering
  \includegraphics[width=\linewidth]{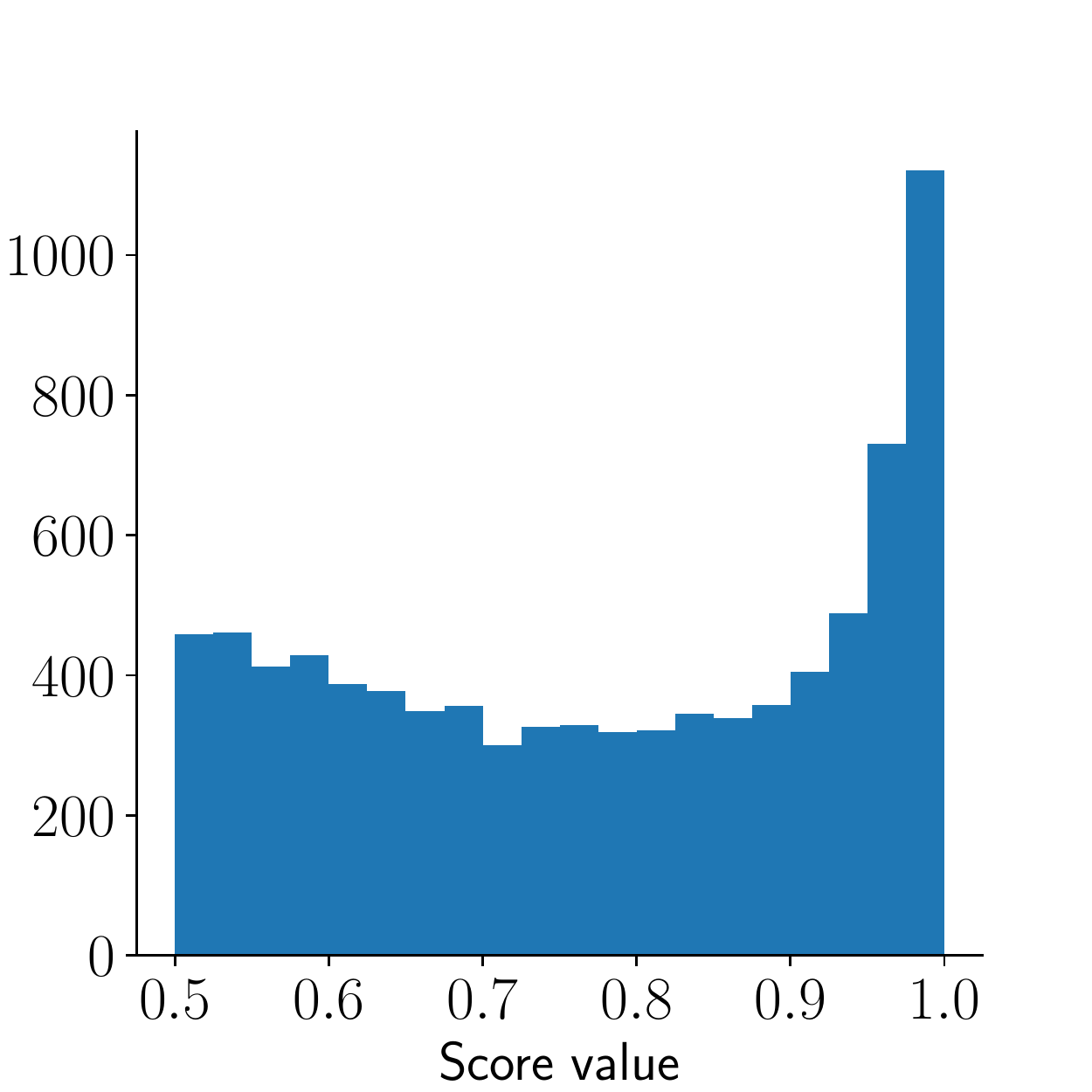}
  \caption{Score histogram for a single class.}
  \label{fig:histo_31}
\end{subfigure}%
\hfill
\begin{subfigure}{.49\linewidth}
  \centering
  \includegraphics[width=\linewidth]{figures/histogram.pdf}
  \caption{Score histogram for all classes.}
  \label{fig:histo_all}
\end{subfigure}
    \caption{Score histograms for a single class ($\selfTrainingParameter=0.7$) (a), and for all the classes ($\selfTrainingParameter=0.75$) (b).} 
    \label{fig:histogram}
\end{figure}

\mysection{Ground threshold.} The key component of our method is its ability to determine a suited threshold value without empirical search. The proposed strategy is to compute the score histogram and set the threshold value to the bin with the lowest number of instances. While the number of bins and the score range are parameters of the proposed method, \Figure{histogram_comparison} shows that the shapes of the histograms are the same, that is U-shaped with high density regions for very low and very high scores, and that they do not influence the position of the lowest density bin. Since this heuristic is independent on the data distribution, it can be applied for each class separately, which gives the possibility to have a set of thresholds rather than a single one. However, training with a uniform threshold seems to achieve better results on \textit{COCO-standard}, as can be seen in \Table{ablation_tau}.
Looking at \Figure{histogram}, which shows the score histogram for a single class and for all the classes jointly, we can see that both of them define a U-shape. This is the shape that we expect since a threshold value lower than the ground threshold would lead to more pseudo-labels, with many of them having a high probability to be false positives. Also, if the score threshold was higher than the ground threshold, we would probably create false negatives. But the problem arises for classes that are hard to learn. For those classes, the chosen heuristic can fail. Since most of the predictions for those classes may have a low confidence score, the histogram can be monotonically decreasing, leading to a ground threshold equal to the last bin. There are multiple consequences to that scenario. First, we generate a lot of false negatives in the pseudo-labeled dataset. Then, the student trains on noisy labels, which will again emphasize the problem of classes hard to learn. Taking a uniform threshold sets the threshold value at a lower score than their ground threshold, leading to fewer false negatives.
However, in the \textit{\dior} setup, we observe that this problem is not present, which we explain by the fact that the teacher is able to better learn the different classes. As shown in \Table{ablation_tau}, the student model performs better with a class-wise threshold.

The key advantage of our method is that it eliminates the need for a parameter search to find the threshold value. However, it is interesting to see how it behaves against this parameter search. \Table{param_search} shows the performance of the student model trained with different threshold values and trained with our method. The results depict two interesting behaviors: (1) the optimal threshold from the parameter search does not always gives a better performance, as can be seen on \textit{\dior}, (2) the optimal threshold value with parameter search for two distinct image distributions does not give the same threshold ($0.7$ for \textit{COCO-standard} and $0.8$ for \textit{\dior}).  This emphasizes that a manual sweep is unsuitable for generalization purposes. It is also important to note that we can process our candidate labels using $6$ bins ranging from $0.5$ to $1$. In this setup, the width of a bin is $0.1$, meaning that the bin with the lowest density will match a value that could have been selected with this classical grid search. On \textit{COCO-standard}, we observed that the ground threshold value in this particular setup is $0.7$, which appears to be the optimal value in \Table{ablation_tau}. This result further consolidate that our ground threshold strategy gives a good threshold value at no cost. Although this particular setup would give the best result for the \textit{COCO-standard} setup, we argue that restraining our method to this particular example does not fulfill our idea to be adaptive to any dataset, which is confirmed with the result obtained on \textit{\dior}.

\begin{table}[ht]
\centering
\caption{Performances obtained with a parameter search on the threshold compared to our method. We report the mean over 5
randomly sampled dataset for both setups.}
\label{tab:param_search}
\resizebox{\columnwidth}{!}{%
\begin{tabular}{ccccccc}
\toprule
$\tau$ & $0.5$ & $0.6$ & $0.7$ & $0.8$ & $0.9$ & Ours  \\
\hline
\textit{COCO $10\%$} & $32.81$ & $32.83$ & $\mathbf{33.02}$ & $32.82$ & $32.57 $ & $32.91$ \\
\textit{DIOR} & $52.14$ & $52.29$ & $52.55$ & $52.61$ & $52.49$ & $\mathbf{52.89}$ \\

\bottomrule
\end{tabular}}
\end{table}



\begin{figure}
    \centering
    \includegraphics[width=\linewidth]{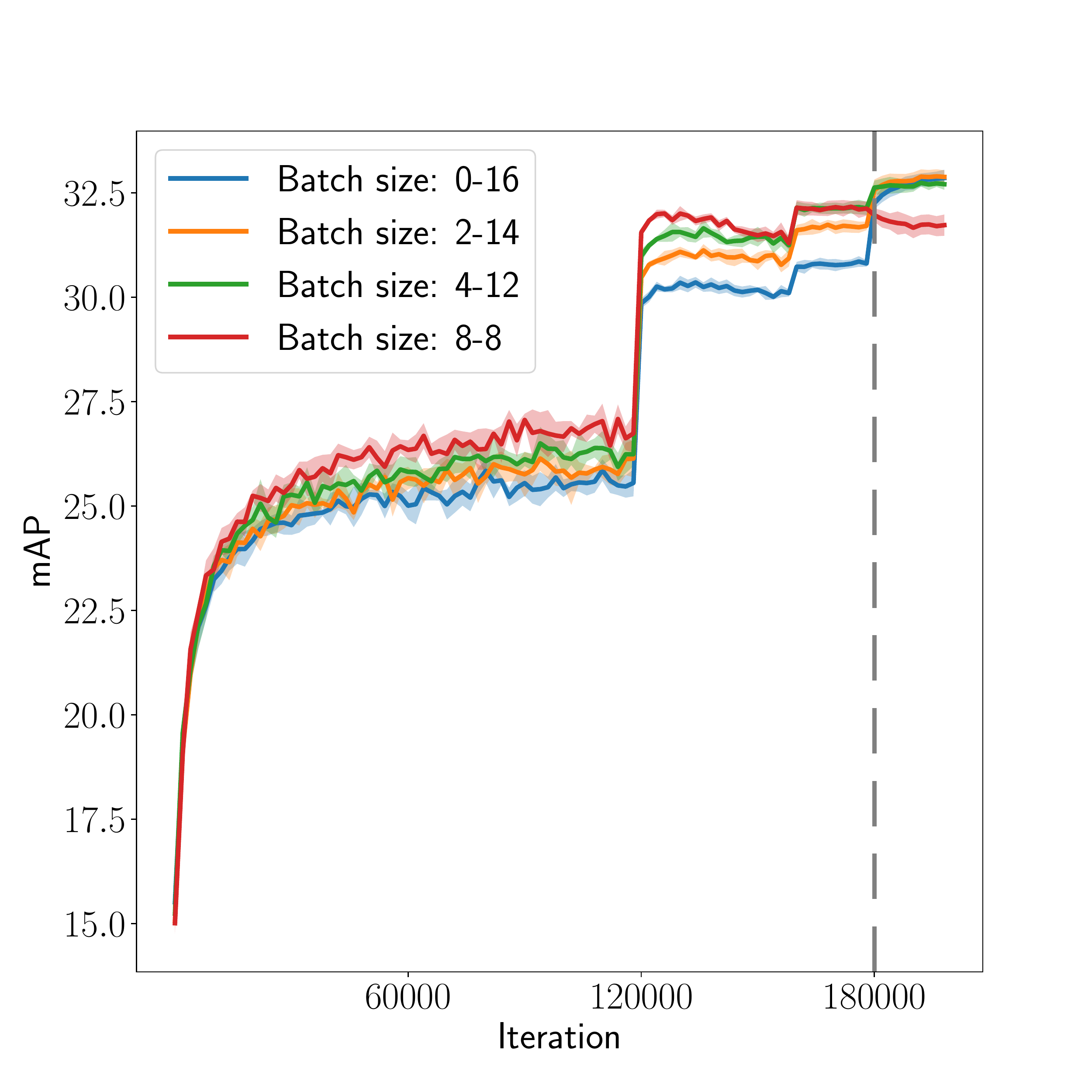}
    \caption{Comparison between the different learning curves of student and refined models \wrt the batch size configuration. The vertical dashed line indicate when the refinement step begins.}
    \label{fig:bs}
\end{figure}

\begin{table}[t]
\centering
\caption{Comparison between refined student models trained with different thresholding strategies. We report the mean and standard deviation over $5$ folds for both setups. Surprisingly, using a uniform value performs better than using one determined class-wise on \textit{COCO-standard}.
    }
\label{tab:ablation_tau}
\begin{tabular}{ccc}
\toprule
 Setup & Class-wise $\selfTrainingParameter$ & Uniform $\selfTrainingParameter$  \\
\hline
\textit{COCO-standard} & $32.56 \pm 0.11$ & $\mathbf{32.91 \pm 0.16}$  \\
\textit{DIOR} & $\mathbf{52.89 \pm 0.33}$ & $52.52 \pm 0.25$ \\

\bottomrule
\end{tabular}
\end{table}

\mysection{Iterative students + Refinement.} Since our method is not based on a mutual training between the teacher and student models, the pseudo-labels used for the student are fixed. However, if we can obtain better results by the end of its training, we can expect to have better candidate labels by using the student as our new teacher. Before we replace the student as our new teacher, we refine the student on the labeled dataset only for a few gradient descent steps. This idea has already been used by Vandeghen~\etal~\cite{Vandeghen2022SemiSupervised} to improve the ROI Heads with only trustable ground-truth labels. As it is shown in \Table{ablation_iteration} for \textit{COCO-standard} and in \Table{ablation_dior} for \textit{DIOR}, this final trick is highly effective. 
The results obtained for $3$ iterations, before and after refinement, are shown in \Table{ablation_iteration}.

During our experiments, we performed an analysis on the batch size distribution between the labeled and pseudo labeled images. The different configurations were $0|16$, $2|14$, $4|12$ and $8|8$, for the labeled and unlabeled size respectively. The averaged learning curves of the first student models are shown in \Figure{bs}, where the training of the student models stops at $180{,}000$ iterations. From the student results, it could be obvious to discard the first two configurations. However, those refined models tend to perform better than the last two configurations. This analysis shows that (1) refining the student models is a crucial step to improve their performance, and (2) drawing some conclusions with only the student performance may not be sufficient.

\begin{table}[t]
\centering
\caption{Comparison between the different iterations of student and refined models for the mAP on \textit{COCO-standard}. There is a twofold message from those results: (1) Consecutive iterations of student training consistently improve the performance compared to the previous iteration. (2) Refining the student is a simple yet effective way to boost the performance. We report the mean and standard deviation over
5 randomly sampled dataset.
    }
\label{tab:ablation_iteration}
\resizebox{\columnwidth}{!}{%
\begin{tabular}{ccccc}
\toprule
  &\multicolumn{1}{c}{$1\%$}     & \multicolumn{1}{c}{$2\%$}    & \multicolumn{1}{c}{$5\%$}    & \multicolumn{1}{c}{$10\%$}     \\ \hline
Supervised & $9.05 \pm 0.16$      &    $12.70 \pm 0.15$    &    $18.47 \pm 0.22$    &    $23.86 \pm 0.81$ \\
Supervised$\dagger$ & $12.14 \pm 0.21$      &    $16.67 \pm 0.30$    &    $23.59 \pm 0.20$    &    $29.34 \pm 0.20$ \\
\hline
Student 1 & $16.57 \pm 0.46$ & $21.53 \pm 0.34$ & $27.64 \pm 0.17$ & $31.77 \pm 0.14$ \\
Refined 1 & $16.67 \pm 0.36$ & $21.93 \pm 0.36$ & $28.47 \pm 0.43$ & $32.91 \pm 0.16$ \\
Student 2 & $17.75 \pm 0.31$ & $23.23 \pm 0.29$ & $29.17 \pm 0.45$ & $32.88 \pm 0.18$ \\
Refined 2 & $17.95 \pm 0.37$ & $23.62 \pm 0.33$ & $29.54 \pm 0.45$ & $33.86 \pm 0.18$ \\
Student 3 & $18.71 \pm 0.30$ & $24.23 \pm 0.34$ & $29.65 \pm 0.41$ & $33.40 \pm 0.23$ \\
Refined 3 & $\mathbf{19.47 \pm 0.39}$ & $\mathbf{24.85 \pm 0.21}$ & $\mathbf{30.43 \pm 0.50}$ & $\mathbf{34.58 \pm 0.22}$ \\
\bottomrule
\end{tabular}}
\end{table}

\section{Conclusion}
\label{sec:Conclusion}

In this paper, we present \methodName, an iterative end-to-end self-training method for object detection. Our method solves the problem of parameter sweep for the threshold value in SSOD with a heuristic threshold value which adapts easily to different setups. We also present the systematic use of a refinement step of the student models to improve their performance. Our experiments show that our method largely outperforms state-of-the-art methods in SSOD, that are threshold-free methods. 

\mysection{Limitations and further work.}
While our method shows an excellent capacity to adapt to diverse data distributions, there is still potential to adapt it to methods which approach the teacher-student scheme with mutual learning. We believe that more work should address the problem of thresholding methods based on parameter search. Finally, a deeper investigation regarding the refining step may be useful, as we have shown that this step consistently improves the performance.

\mysection{Acknowledgments.}
The present research benefited from computational resources made available on Lucia, the Tier-1 supercomputer of the Walloon Region, infrastructure funded by the Walloon Region under the grant agreement n°1910247.

\clearpage

{\small

}

\clearpage
\section{Supplementary Material}

\subsection{Implementation details.}

\mysection{Networks.}
We use a pre-trained ResNet-50~\cite{He2016DeepResidual} as backbone for Faster-RCNN~\cite{Ren2017Faster} with FPN~\cite{Lin2017Feature} as object detector.

\mysection{Training parameters.}
For the \textit{COCO-standard} setup, the teacher models are warmed-up for $20$k steps with a learning rate decay after $13$k and $18$k steps. Then, our student models are trained for $180$k steps, using a global batch size of $64$. We apply the same learning rate decay after $120$k and $160$k steps. We use SGD as optimizer, with an initial learning rate of $0.08$ and with default other parameters. The refined student models are trained for $10$k steps, using a initial learning rate of $0.0008$, which is reduced after $6$k steps.
For the \textit{DIOR} setup, the student models are trained for $90$k steps with a learning rate decay after $60$k and $80$k steps. The different values are gathered in \Table{parameters}.

\begin{table}[!h]
\centering
\caption{Hyper-parameters used during the training of the different models}
\label{tab:parameters}
\resizebox{\columnwidth}{!}{%
\begin{tabular}{ccccccc}
\toprule
Parameters & \multicolumn{3}{c}{{\textit{COCO-standard}}} & \multicolumn{3}{c}{{\textit{DIOR}}} \\
 & Teacher & Student & Refined & Teacher & Student & Refined \\
 \hline
Training steps & $20$k & $180$k & $10$k & $20$k & $90$k & $10k$ \\
Learning rate & $0.08$ & $0.08$ & $0.08$ & $0.08$ & $0.08$ & $0.08$ \\
Learning rate decay & $13$k-$18$k & $120$k-$160$k & $6$k & $13$k-$18$k & $60$k-$80$k & $6$k \\
Batch Size (labeled | pseudo labeled) & $64|0$ & $8|56$ & $64|0$ & $32|0$ & $8|24$ & $32|0$ \\
\bottomrule
\end{tabular}}
\end{table}

\mysection{Data augmentations.}
For the data augmentations during training, we use some large scale color jittering, such as random changes in brightness, contrast, hue and saturation. We also apply some scale jittering and random horizontal flips.

\subsection{Student training.}

In \Table{ablation_pseudo}, we show the results of refined student models trained with pseudo-labels generated with different view strategies. The idea of using the four different views (normal, flip, scale and flip+scale). We can see that only scaling up the view gives worse results, but scaling and flipping gives a tiny improvement compared to only flipping the image.

\begin{table}[!ht]
\centering
\caption{Comparison between refined student models trained with different view techniques during the generation of candidate labels for the mAP on \textit{COCO-standard}. For the Scale+Flip technique, we use the information of the normal view, the scaled/flipped only view and the scale+flip view. Adding multiple views is a simple yet effective way to improve the quality of candidate labels. We report the mean and standard deviation over 5 randomly sampled dataset.}
\label{tab:ablation_pseudo}
\resizebox{\columnwidth}{!}{%
\begin{tabular}{ccccc}
\toprule
Transformation & None & Scale & Flip & Scale+Flip \\
\hline
mAP & $32.87 \pm 0.23$ & $32.76 \pm 0.17$ & $32.90 \pm 0.19$ & $\mathbf{32.91 \pm 0.16}$ \\

\bottomrule
\end{tabular}}
\end{table}

We also study the effect of weighting the loss in \Equation{loss_u} during the training of the student. We trained a student by fixing the $\alpha$ term in \Equation{loss_u} to $1$. On average, the gain of mAP is $0.13$ for the model trained with the weighted loss.

\end{document}